\pdfoutput=1
\documentclass[11pt]{article}

\usepackage[]{emnlp2021}

\usepackage{times}
\usepackage{latexsym}

\usepackage[T1]{fontenc}
\usepackage[utf8]{inputenc}

\usepackage{microtype}

\usepackage{makecell}

\usepackage{hyperref}
\usepackage{xcolor,colortbl}

\definecolor{Gray}{gray}{0.90}
\newcolumntype{g}{>{\columncolor{Gray}}c}

\usepackage{latexsym}
\usepackage{url}
\usepackage{soul}
\usepackage{graphicx}  %Required

\usepackage{subcaption}

\usepackage{amsmath} 
\usepackage{amssymb} 
\usepackage{paralist}
\usepackage{booktabs}
\usepackage{algorithm}
\usepackage{algorithmic}
\usepackage{multirow}
\usepackage{diagbox}

\usepackage{calc}  
\usepackage{enumitem}  
\title{Variation and generality in encoding of syntactic anomaly information in sentence embeddings}

\author{Qinxuan Wu\thanks{{ }This work was done while the first author was a visiting student at the University of Chicago.} \\
  College of Computer Science and Technology \\
  Zhejiang University \\
  \texttt{wuqinxuan@zju.edu.cn} \\\And
  Allyson Ettinger \\
  Department of Linguistics \\
  The University of Chicago \\
  \texttt{aettinger@uchicago.edu} \\}

\begin{document}
\maketitle

\begin{abstract}
While sentence anomalies have been applied periodically for testing in NLP, we have yet to establish a picture of the precise status of anomaly information in representations from NLP models. In this paper we aim to fill two primary gaps, focusing on the domain of syntactic anomalies. First, we explore fine-grained differences in anomaly encoding by designing probing tasks that vary the hierarchical level at which anomalies occur in a sentence. Second, we test not only models' ability to detect a given anomaly, but also the generality of the detected anomaly signal, by examining transfer between distinct anomaly types. Results suggest that all models encode some information supporting anomaly detection, but detection performance varies between anomalies, and only representations from more recent transformer models show signs of generalized knowledge of anomalies. Follow-up analyses support the notion that these models pick up on a legitimate, general notion of sentence oddity, while coarser-grained word position information is likely also a contributor to the observed anomaly detection.
\end{abstract}

\section{Introduction}

As the NLP community works to understand what is being learned and represented by current models, a notion that has made sporadic appearances is that of linguistic anomaly.
% \cite{Conneau18,encoder_grammatical_error,pham2020out}. 
Analyses of language models have often tested whether models prefer grammatical over ungrammatical completions~\citep[e.g.][]{Linzen}, while analyses of sentence embeddings have probed for syntax and semantics by testing detection of sentence perturbations~\citep{Conneau18}. Such work tends to exploit anomaly detection as a means of studying linguistic phenomena, setting aside any direct questions about encoding of anomaly per se. However, models' treatment of anomaly is itself a topic that raises important questions. After all, it is not obvious that we should expect models to encode information like ``this sentence contains an anomaly'', nor is it obvious which types of anomalies we might expect models to pick up on more or less easily. Nonetheless, anomalies are easy to detect for humans, and their detection is relevant for applications such as automatic error correction \cite{ge2018reaching}, so it is of value to understand how anomalies operate in our models, and what impacts anomaly encoding. 

In the present work we seek to fill this gap with a direct examination of anomaly encoding in sentence embeddings. We begin with fine-grained testing of the impact of anomaly type, designing probing tasks with anomalies at different levels of syntactic hierarchy to examine whether model representations better support detection of certain types of anomaly. Then we examine the generality of anomaly encoding by testing transfer performance between distinct anomalies---here our question is, to the extent that we see successful anomaly detection, does this reflect encoding of a more general signal indicating ``this sentence contains an anomaly'', or does it reflect encoding of simpler cues specific to a given anomaly? We focus on syntactic anomalies because the hierarchy of sentence structure is conducive to our fine-grained anomaly variation. (Sensitivity to syntactic anomalies has also been studied extensively as part of the human language capacity~\cite{chomsky1957syntactic,fodor1996tasks}, strengthening precedent for prioritizing it.)

We apply these tests to six prominent sentence encoders. We find that most models support non-trivial anomaly detection, though there is substantial variation between encoders. We also observe differences between hierarchical classes of anomaly for some encoders.
When we test for transferability of the anomaly signal, we find that for most encoders the observed anomaly detection shows little sign of generality---however, transfer performance in BERT and RoBERTa suggests that these more recent models may in fact pick up on a generalized awareness of syntactic anomalies. Follow-up analyses support the possibility that these transformer-based models pick up on a legitimate, general notion of syntactic oddity---which appears to coexist with coarser-grained, anomaly-specific word order cues that also contribute to detection performance. We make all data and code available for further testing.\footnote{ \url{https://github.com/pepper-w/syntactic\_anomalies}.}

\section{Related Work}\label{relaed_work}

This paper builds on work analyzing linguistic knowledge reflected in representations and outputs of NLP models \cite{tenney2019you,rogers2020primer,jawahar2019does}. Some work uses tailored challenge sets associated with downstream tasks to test linguistic knowledge and robustness~\cite{XYZ,NMT->NLI,DNC,DNC17,diff_layer, yang2015wikiqa, rajpurkar2016squad, jia2017adversarial, rajpurkar2018know}.
Other work has used targeted classification-based probing to examine encoding of specific types of linguistic information in sentence embeddings more directly~\cite{Adi2016,Conneau18,mophological,ettinger2016probing,ettinger2018assessing,tenney2019you,klafka2020spying}. 
We expand on this work by designing analyses to shed light on encoding of syntactic anomaly information in sentence embeddings.

A growing body of work has examined syntactic sensitivity in language model outputs
\cite{syntac_chowdhury2018rnn,syntac_futrell2019neural,syntac_lakretz2019emergence,syntac_marvin2018targeted, ettinger2020bert}, and our \emph{Agree-Shift} task takes inspiration from the popular number agreement task for language models~\cite{Linzen, colorless_green,goldberg2019assessing}. 
Like this work, we focus on syntax in designing our tests, but we differ from this work in focusing on model representations rather than outputs, and in our specific focus on understanding how models encode information about anomalies. Furthermore, as we detail below, our \emph{Agree-Shift} task differs importantly from the LM number agreement tests, and should not be compared directly to results from those tests.

Our work relates most closely to studies involving anomalous or erroneous sentence information
\cite{cola,encoder_grammatical_error,parser_robustness}.
Some work investigates impacts from random shuffling or other types of distortion of input text \citep{pham2020out,gupta2021bert} or of model pre-training text \cite{sinha2021masked} on downstream tasks---but this work does not investigate models' encoding of these anomalies.~\citet{cola} present and test with the \textit{CoLA} dataset for general acceptability detection, and among the probing tasks of~\citet{Conneau18} there are three that involve analyzing whether sentence embeddings can distinguish erroneous modification to sentence inputs: \textit{SOMO}, \textit{BShift}, and \textit{CoordInv}.
\citet{encoder_grammatical_error} 
also generate synthetic errors based on errors from non-native speakers, 
showing impacts of such errors on downstream tasks, and briefly probing error sensitivity. 
More recently, \citet{li2021bertsurprised} conduct anomaly detection with various anomaly types at different layers of transformer models, using training of Gaussian models for density estimation, and finding different types of anomaly sensitivity at different layers.
We build on this line of work in anomaly detection with a fine-grained exploration of models' detection of word-content-controlled perturbations at different levels of syntactic hierarchy. Our work is complementary also in exploring generality of models' anomaly encoding by examining transfer performance between anomalies.

\section{Syntactic Anomaly Probing Tasks}~\label{section3}
To test the effects of hierarchical location of a syntactic anomaly, we create a set of tasks based on four different levels of sentence perturbation.
We structure all perturbations so as to keep word content constant between original and perturbed sentences, thus removing any potential aid from purely lexical contrast cues.
Our first three tasks involve reordering of syntactic constituents, and differ in hierarchical proximity
of the reordered constituents: the first switches constituents of a noun phrase, the second switches constituents of a verb phrase, and the third switches constituents that only share the clause. Our fourth task tests sensitivity to perturbation of morphological number agreement, echoing existing work testing agreement in language models~\cite{Linzen}.

\subsection{\emph{Mod-Noun}: Detecting modifier/noun reordering}
Our first task tests sensitivity to anomalies in modifier-noun structure, generating anomalous sentences by swapping the positions of nouns and their accompanying modifiers, as below:

% \begin{description}[leftmargin=!,labelwidth=\widthof{\bfseries The longest label}]
\begin{description}[leftmargin=!,labelwidth=0pt]
\item[] 
\textit{A man wearing a \textbf{\underline{yellow} scarf} rides a bike. $\rightarrow$
\\A man wearing a \textbf{scarf \underline{yellow}} rides a bike.}
\end{description}

We call this perturbation \textit{Mod-Noun}.  
Any article determiner of the noun phrase remains unperturbed.

\subsection{\emph{Verb-Ob}: Detecting verb/object reordering}
Our second task tests sensitivity to anomalies in English subject-verb-object (SVO) sentence structure by swapping the positions of verbs and their objects (SVO $\rightarrow$ SOV). To generate perturbed sentences for this task, we take sentences with a subject-verb-object construction, and reorder the verb (or verb phrase) and the object, as in the example below:

% \begin{description}[leftmargin=!,labelwidth=\widthof{\bfseries The longest label}]
\begin{description}[leftmargin=!,labelwidth=0pt]
\item[] \textit{A man wearing a yellow scarf \textbf{\underline{rides} a bike}. $\rightarrow$ 
 \\A man wearing a yellow scarf \textbf{a bike \underline{rides}}.}
\end{description}

We refer to this perturbation as \textit{Verb-Ob}. 
Note that \textit{Verb-Ob} and \textit{Mod-Noun} are superficially similar tasks in that they both reorder sequentially consecutive constituents. However, importantly, they differ in the hierarchical level of the swap. 

\subsection{\emph{SubN-ObN}: Detecting subject/object reordering} 
Our third task tests sensitivity to anomalies in subject-verb-object relationships, creating perturbations by swapping the positions of subject and object nouns in a sentence. 
For this task, we generate the data by swapping the two head nouns of the subject and the object, as below:
% by selecting sentences with constructions of ``NP + VP + NP'', 
% The task of across-constituent reorder: \textit{SubN-ObN}. 
% (NP1(NN1) + VP + NP2(NN2) $\rightarrow$ NP1(NN2) + VP + NP2(NN1)).

% \begin{description}[leftmargin=!,labelwidth=\widthof{\bfseries The longest label}]
\begin{description}[leftmargin=!,labelwidth=0pt]
\item[] \textit{A \textbf{\underline{man}} wearing a yellow scarf rides a \textbf{bike}. $\rightarrow$
 \\A \textbf{bike} wearing a yellow scarf rides a \textbf{\underline{man}}.}
\end{description}

We refer to this perturbation as \textit{SubN-ObN}. 
We target subject-verb-object structures directly under the root of the syntactic parse, meaning that only one modification is made per sentence for this task.

Detecting the anomaly in this perturbation involves sensitivity to argument structure (the way in which  subject, verb, and object should be combined), along with an element of world knowledge (knowing that a bike would not ride a man, nor would a bike typically wear a scarf).\footnote{For more details about this task, please see Appendix A.7.}

\subsection{\emph{Agree-Shift}: Detecting subject/verb disagreement}
Our fourth task tests sensitivity to anomalies in subject-verb morphological agreement, by changing inflection on a present tense verb to create number disagreement between subject and verb:

% \begin{description}[leftmargin=!,labelwidth=\widthof{\bfseries The longest label}]
\begin{description}[leftmargin=!,labelwidth=0pt]
\item[] \textit{\textbf{A man} wearing a yellow scarf \textbf{\underline{rides}} a bike. $\rightarrow$ 
 \\\textbf{A man} wearing a yellow scarf \textbf{\underline{ride}} a bike.}
\end{description}

We refer to this perturbation as \textit{Agree-Shift}.\footnote{Note that while this perturbation echoes the popular LM agreement analyses~\cite{Linzen,colorless_green,goldberg2019assessing}, the fact that we are probing sentence embeddings for explicit detection of this anomaly is an important difference. 
Performance by LMs on those agreement tasks can indicate that a model prefers a non-anomalous completion, but cannot speak to whether the model encodes any explicit/perceptible awareness that an anomaly is present/absent. For this reason, model performance on our \emph{Agree-Shift} task should not be compared directly to performance on these agreement probability tasks.}\label{fn}
This is the only one of our tasks that involves a slight change in the word inflection, but the word stem remains the same---we consider this to be consistent with holding word content constant.

\section{Experiments}

We generate probing datasets for each of the anomaly tests described above. We then apply these tasks to examine anomaly sensitivity in a number of generic sentence encoders.\footnote{Please refer to the Appendix A.5, A.6, A.7 for more details about data generation, probing implementation, as well as descriptions about encoders and external tasks.}

\paragraph{Datasets}\label{sec:datasets}

Each of the above perturbations is used to create a probing dataset consisting of normal sentences and corresponding modified sentences, labeled as \textit{normal} 
and \textit{perturbed}, respectively.
Within each probing task, each \emph{normal} sentence has a corresponding \emph{perturbed} sentence, so the label sets for each task are fully balanced.
Each probe is formulated as a binary classification task. Normal sentences and their corresponding perturbed sentences are included in the same partition of the train/dev/test split, so for any sentence in the test set, no version of that sentence (neither the perturbed form nor its original form) has been seen at training time.
We draw our \textit{normal} sentences from MultiNLI~\cite{multiNLI} (premise only).
Perturbations of those sentences are then generated as our \textit{perturbed} sentences.

\paragraph{Probing}~\label{encoders}
We analyze sentence embeddings from these prominent sentence encoders:
InferSent~\cite{infersent}, Skip-thoughts~\cite{skipthoughts}, GenSen~\cite{gensen}, BERT~\cite{bert}, and RoBERTa \cite{liu2019RoBERTa}.
The first three models are RNN-based, while the final two are transformer-based.

To test the effectiveness of our control of word content, we also test bag-of-words (BoW) sentence embeddings obtained by averaging of
% simple averaging of word embeddings produced by 
GloVe~\cite{glove} embeddings. This allows us to verify that our probing tasks are not solvable by simple lexical cues~\cite[c.f.][]{ettinger2018assessing}, thus better isolating effects of syntactic anomalies.

We train and test classifiers on our probing tasks, with sentence embeddings from the above encoders as input. The classifier structure is a multilayer perceptron (MLP) classifier with one hidden layer.\footnote{We also train on a logistic regression (LR) classifier. LR results are shown in the Appendix Table \ref{tab:LR-results}.}

\section{Anomaly Detection Results}\label{experiment_one}

\begin{figure}[t!]
 \centering
 \includegraphics[width=0.48\textwidth]{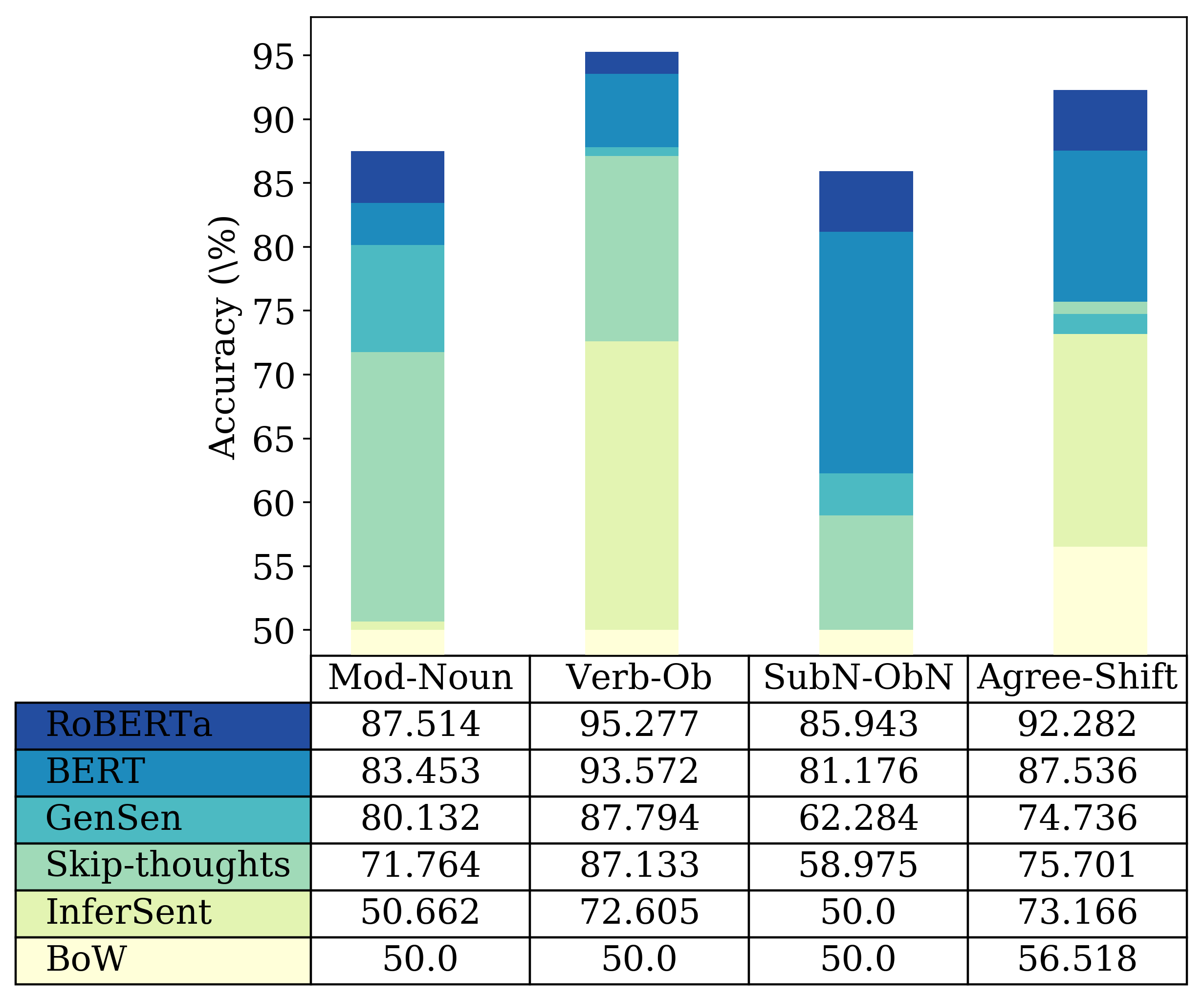}
 \caption{Anomaly detection performance.}
 \label{fig:original_task_only}
\end{figure}

Fig.~\ref{fig:original_task_only} 
% \ake{the figure is small---if possible we should enlarge it (e.g., put it across two columns)}
shows anomaly detection performance for the tested encoders.
% \ake{we do need to do something about the grayscale---too difficult to identify which model is which}.
We can see first that for three of our four tasks---all reordering tasks---our BoW baseline performs perfectly at chance, verifying elimination of lexical biases. BoW on the \textit{Agree-Shift} task is just above chance, reflecting (expected) slight bias in morphological variations.

% Looking at patterns of performance between our generated probing tasks, 
Comparing between tasks,
we see that \textit{Verb-Ob} yields highest overall performance while \textit{SubN-ObN} yields the lowest. 
% In particular, \textit{Verb-Ob} represents one of the highest syntactic levels at which a constituent swap occurs (the level of the verb phrase). 
As mentioned above, a particularly informative comparison is between \textit{Verb-Ob} and \textit{Mod-Noun}, which both involve swapping sequentially adjacent content, but at different hierarchical levels. We see that encoders consistently show stronger performance on \textit{Verb-Ob} than \textit{Mod-Noun}, suggesting that the broader hierarchical domain of  \textit{Verb-Ob} may indeed make anomalies more accessible for encoding. The only anomaly that affects a broader span of the sentence is \textit{SubN-ObN}---but we see that this is instead one of the most challenging tasks. We suspect that this is attributable to the fact that, as described above, detecting this anomaly may require extra world knowledge and common sense, which certain encoders may have less access to.\footnote{As an encoder trained on semantic reasoning, InferSent nonetheless fails terribly on this task---this may be explained by findings that heuristics can account for much of NLI task learning~\cite{poliak2018hypothesis,mccoy2020right}.} It is not unexpected, then, that BERT and RoBERTa, with comparatively much larger and more diverse training data exposure, show a large margin of advantage on this challenging \textit{SubN-ObN} task relative to the other encoders. \textit{Agree-Shift} patterns roughly on par with \textit{Mod-Noun}, though notably InferSent (and Skip-thoughts) detects the agreement anomaly much more readily than it does the \textit{Mod-Noun} anomaly.
% in part to the combination of information sources relevant for detection of \textit{SubN-ObN} anomalies: 

Comparing between encoders, we see clear stratification in performance. InferSent shows the least anomaly awareness, performing for half of the tasks at chance level with the BoW baseline. GenSen and Skip-thoughts, by contrast, consistently occupy a higher performance tier, often falling not far behind (but never quite on par with) the highest level of performance. The latter distinction is reserved for BERT and RoBERTa, which show the strongest anomaly sensitivity on all tasks. 
All models show stronger performance on \emph{Verb-Ob} than \emph{Mod-Noun}, but the hierarchical difference between these tasks seems to have particularly significant impact for InferSent and Skip-thoughts, with cues relating to \emph{Verb-Ob} seemingly encoded by InferSent, but cues relating to \emph{Mod-Noun} seeming to be absent. \textit{Mod-Noun} also yields the largest margin of difference between GenSen and Skip-thoughts. Since Skip-thoughts is one objective of GenSen, this suggests that the additional GenSen objectives provide an edge particularly for the finer-grained information needed for \emph{Mod-Noun}.

BERT and RoBERTa emerge soundly as the strongest encoders of anomaly information, with RoBERTa also consistently outperforming BERT.
While this is in line with patterns of downstream task performance from these models, it is noteworthy that these models also show superior performance on these anomaly-detection tasks, as it is not obvious that encoding anomaly information would be relevant for these models' pre-training objectives, or for the most common NLU tasks on which they are typically evaluated.

\section{Investigation on Generality of Anomaly Encoding}\label{sec:transfer}

The above experiments suggest that embeddings from many of these encoders contain signal enabling detection of the presence of syntactic anomalies. However, these results cannot tell us whether these embeddings encode awareness that an ``error'' is present per se---the classifiers may simply have learned to detect properties associated with a given anomaly, e.g., agreement mismatches, or occurrence of modifiers after nouns. In this sense, the above experiments serve as finer-grained tests of levels of hierarchical information available in these embeddings, but still do not test awareness of the notion of anomaly in general.

% It’s less obvious that encoders need this type of information for language understanding. However, we motivate probing for this kind of general anomaly awareness by pointing out that humans possess it \cite{fodor1996tasks,chomsky1957syntactic}.
In this section we take a closer look at anomaly awareness per se, by testing the extent to which the sensitivities identified above are specific to individual anomalies, or reflective of a more abstract ``error'' signal that would apply across anomalies. We explore this question by testing transfer performance between different anomaly types.

\begin{figure*}[t!]
%  \centering
\flushright
\includegraphics[width=0.85\textwidth]{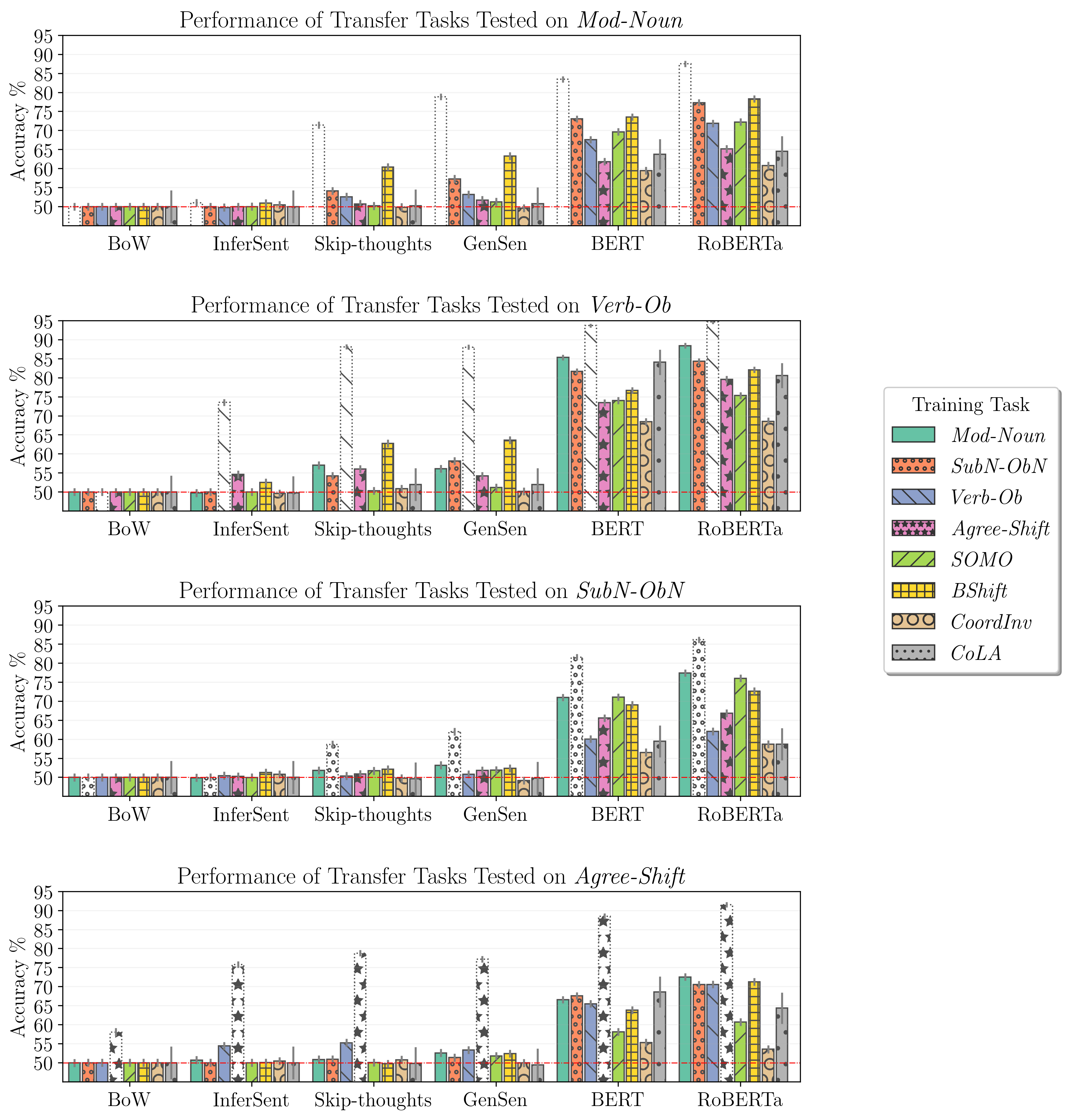}
 \caption{Results on transfer settings, by test tasks. 
 Different patterns/colors represent different training tasks. 
%  Different colors demonstrate different encoders. 
 Results for non-transfer settings (same training and test tasks) are shown in white bars.
%  All the error bars represent 95\% binomial confidence intervals. Dotted horizontal line indicates chance-level accuracy. 
 }
 \label{fig:multi_task}
\end{figure*}

% \subsection{Transfer Results}
\paragraph{Transfer Results}

While in Section~\ref{experiment_one} we focused on examining anomaly-specific sensitivity in our new tasks---testing variation along fine-grained syntactic hierarchical distinctions and in a word-controlled setting---for examining generality of anomaly encoding it is worthwhile to take into account a broader range of anomaly types and datasets.
For this reason we examine transfer between each of our generated probing tasks, as well as transfer to our tasks from established datasets: \textit{SOMO}, \textit{BShift}, and \textit{CoordInv} from \citet{Conneau18}, and the \textit{CoLA} task~\cite{cola}.

Fig.~\ref{fig:multi_task}
% \ake{those figures are much too small right now, so we'll need to think about how to fix that}
shows transfer results from each dataset to each of our tasks.
% \footnote{
% Transparent bars show the results on LR classifier in the case that LR yields better accuracy than MLP (non-transparent). \ake{check whether we decide to keep LR}}
% (with bar hatching identifying the training anomaly). 
For ease of comparison, we also show the test result achieved when training on the same anomaly (the non-transfer result) in the white bars.
% \ake{It might be good to include in Fig.~\ref{fig:multi_task} the bars that represent train/test on the same task, for the sake of comparison (and see my above comment about potentially using the same test set as the first experiments). I'm thinking this bar should always go on the far left or far right of the cluster, so it's easy to identify.}
We see that the majority of encoders show a marked drop in performance relative to the original non-transfer accuracy, and in fact most of the transfer results are approximately at chance performance. This suggests that the embeddings from these models encode information supporting detection of these anomalies, but the signals that enable this detection are anomaly-specific. That is, they may encode some syntactic/semantic signal supporting detection of specific anomalies, but there is no indication that they encode a general awareness that ``there is an anomaly''. 

The notable exceptions to this poor transfer performance are the transformer-based models, BERT and RoBERTa, which by stark contrast to the RNN-based encoders, show non-trivial transfer performance across all four of our generated tasks, regardless of the anomaly that the classifier is trained on.
This suggests that to a much greater extent than any of the other encoders, BERT and RoBERTa may encode a more general ``error'' signal, allowing for generalization across anomalies. Importantly, BERT and RoBERTa do also show some performance drop from non-transfer to transfer settings---so while they may encode a more generalized ``error'' signal, this is likely in combination with encoding of anomaly-specific information that further aids performance on a given anomaly.

We note that transfer performance from \textit{CoLA} is typically comparable to, or occasionally better than, training on the Conneau et al. tasks--despite the fact that \textit{CoLA} has much smaller training data. \emph{CoLA} is also the only task that contains a variety of anomalies for the model to learn from, rather than a single anomaly type as in all other datasets. This may enable faster, more generalizable learning on this small dataset---but of course, this would only be possible if a generalized anomaly signal is available in the embeddings. Following this reasoning, we also test whether jointly training on multiple anomalies improves transfer performance. The results of these multi-task transfer experiments can be found in Appendix Tables \ref{tab:multi_task_new_test_on_our_task_cosis_size}-\ref{tab:multi_task_new_conneau_generated_cosis_size}. These transfer results show overall a small decrease in performance relative to the one-to-one transfer, suggesting that training on single types of anomalies is not creating any major disadvantage for transfer performance. It may also indicate that mixed types of higher-level oddity in natural occurring anomalies from \textit{CoLA} is not trivial to simulate by stacking together data with single type of anomalies as we do here.

The Conneau et al.~task that most often shows the best transfer to our tasks (especially \textit{Mod-Noun} and \textit{Verb-Ob}) is \textit{BShift}. This is sensible, given that that task involves detecting a switch in word order within a bigram. Given this similarity, we can expect to see some transfer from this task even in the absence of generalized anomaly encoding.

As for how our generated anomaly types vary in supporting transfer to other anomaly types, we again note some differences between \textit{Mod-Noun} and \textit{Verb-Ob}. While \textit{Verb-Ob} proved more accessible for detection than \textit{Mod-Noun}, in the transfer setting we find that the broader hierarchical perturbation in \textit{Verb-Ob} is often less conducive to transfer than \textit{Mod-Noun}. Below we explore further to better understand what models are learning when trained on these anomalies.

\section{Further analyses}

\subsection{Exploring false positives}

The results above indicate that embeddings from these encoders contain non-trivial signal relevant for specific perturbations, but only BERT and RoBERTa show promise for encoding more general awareness of anomalies. To further explore the anomaly signal learned from these representations, in this section we apply the learned anomaly classifiers to an entirely new dataset, for which no perturbations have been made. For this purpose, we use the \textit{Subj-Num} dataset from~\citet{Conneau18}.\footnote{Dataset size is 10k sentences (test set).} By default we can assume that all sentences in these data are non-anomalous, so any sentences labeled as \textit{perturbed} can be considered errors. After testing the pre-trained classifiers on embeddings of these unperturbed sentences, we examine these false positives to shed further light on what the classifiers have come to consider anomalies. We focus this analysis on BERT and RoBERTa, as the two models that show the best anomaly detection and the only signs of generalized anomaly encoding.

\begin{table}[t]
\centering
\resizebox{0.45\textwidth}{!}{%
\begin{tabular}{l|l|l|l|l}
\toprule
        %   & \multicolumn{4}{c|}{original}                                                   \\ \midrule
Train task & \textit{Mod-Noun} & \textit{Verb-Ob} & \textit{SubN-ObN} & \textit{Agree-Shift} \\ \hline
BERT       & 4.9\%             & 2.61\%           & 4.81\%            & 31.98\%               \\ \hline
RoBERTa    & 7.73\%            & 3.6\%            & 7.83\%           & 22.13\%              \\ \toprule
%           & \multicolumn{4}{c|}{non-syntactic}                                              \\ \midrule
% train task & \textit{Mod-Noun} & \textit{Verb-Ob} & \textit{SubN-ObN} & \textit{Agree-Shift} \\ \hline
% BERT       & 1.82\%            & 3.75\%           & 2.0\%             & 22.86\%              \\ \hline
% RoBERTa    & 0.96\%            & 1.64\%           & 1.54\%            & 4.92\%               \\ \bottomrule
\end{tabular}%
}
\caption{Error rate on \textit{Subj-Num} data. 
% The top rows show the results when the training on original tasks, while the bottom rows show the results when training on non-syntactic tasks.
% \blue{TODO: move to appendix}
}\label{tab:wrongly_labeled_rate}
\end{table}

Error rates for this experiment are shown in Appendix Table~\ref{tab:wrongly_labeled_rate}. We see that in general the false positive rates for these models are very low.
% ---lower than the original probing error rates for both models.
% , and on par with the original ones for GenSen. 
The highest error rates are found for the classifiers trained on \emph{Agree-Shift}, and examination of these false positives suggests that the majority of these errors are driven by confusion in the face of past-tense verbs (past tense verbs do not inflect for number in English---so past tense verbs were few, and uninformative when present, in our \emph{Agree-Shift} task). This type of error is less informative for our purposes, so we exclude the \emph{Agree-Shift} classifier for these error analyses. For the other classifiers, the error rates are very low, suggesting that the signal picked up on by these classifiers is precise enough to minimize false positives in \textit{normal} inputs.
% \footnote{Agree-Shift, though with higher error rates, most of the sentences with spurious positive labels are with past tense verb, which is the unseen case for the Agree-Shift training data. Albeit unseen cases given with positive labels or labeled randomly (ideally) is a reasonable behavior for classifiers, this might also indicate some potential limitation of this task to be broader applicable to natural occurring data.}

\begin{table*}[t!]
\centering
\resizebox{\textwidth}{!}{%
\normalsize
\begin{tabular}{@{}l|l@{}}
\toprule

% \\ \midrule
        % & \multicolumn{1}{c|}{\cellcolor[HTML]{C0C0C0}Intersection between three reordering tasks}                                                                                             
%         \\ \midrule
% GenSen  & \begin{tabular}[c]{@{}l@{}}A gale of laughter rippled through the viewing stands .\\ The only guarantee I had that the boss was inside came from the bear Casey had dated , Fin .\end{tabular}      
% \\ \midrule
BERT    & \begin{tabular}[c]{@{}l@{}}
\makecell[l]{Dolores asked pointing to a sway backed building made in part of logs and cover with a tin roof .}\\
Fireworks$^\ast$ , animals woven of fire and women dancing with flames .\\
There were nice accessible veins there .\\
Three rusty screws down and Beth spoke , making him jump .\\
The pillars were still warming up , but the more powerful they got the more obvious it became to Mac about what was going on .\\
The signals grew clearer , voices , at first faint , then very clear .\\
\end{tabular}                                                       \\ \midrule
RoBERTa & \begin{tabular}[c]{@{}l@{}}
% Recycle , \textbf{``} replied the students altogether .
\makecell[l]{One row , all the way across , formed words connected without spaces .}
\\
And kidnappers with God only knew what agenda . \\
% \textbf{`` }Cousin ,\textbf{ ``} Innocent said , \textbf{``} perhaps your natural philosopher has the right of it .
The slums would burn , not stone nobleman keeps . ``\\
`` Hull reinforcements are out of power .\\
From inside that pyramid seventy centuries look out at us .\\
The ornaments$^\ast$ she wore sparkled but isn 't noticeable much , as her blissful countenance shined over , surpassing it .\\
\end{tabular}  
\\ \bottomrule
\end{tabular}%
}
\caption{Representative false positives shared by all three reordering classifiers.}\label{tab:error_analysis_intersect}
\end{table*}

To examine generality of the anomaly signal detected by the classifiers, we look first to sentences that receive false positives from multiple of the three reordering-based classifiers. We find that within the union of false positives identified across all three classifiers, sentences that are labeled as anomalous by at least two classifiers make up 28.6\% and 35.6\% for BERT and RoBERTa respectively---and sentences labeled as anomalous by all three classifiers make up 7.3\% and 9.6\%. Since no two classifiers were trained on the same perturbation, the existence of such overlap is consistent with some generality in the anomaly signal for the representations from these two models.
% ---with RoBERTa encoding showing signs of slightly more generality.

Table~\ref{tab:error_analysis_intersect} lists samples of false positives identified by all three classifiers. While these sentences are generally grammatical, we see that many of them use somewhat convoluted structures---in many cases one can imagine a human requiring a second pass to parse these correctly. In some cases, as in the ``Fireworks'' example, there is not a full sentence---or in the ``ornaments'' example, there appears to be an actual ungrammaticality. The fact that the classifiers converge on sentences that do contain some structural oddity supports the notion that these classifiers may, on the basis of these models' embeddings, have picked up on somewhat of a legitimate concept of syntactic anomaly.

Of course, there are also many items that individual classifiers identify uniquely. We show examples of these in Appendix Table~\ref{tab:error_analysis_exclusive}. The presence of such anomaly-specific errors is consistent with our findings in Section~\ref{sec:transfer}, that even with BERT and RoBERTa the classifiers appear to benefit from some anomaly-specific signal in addition to the potential generalized anomaly signal. 

Examining these classifier-specific false positives, we can see some patterns emerging. The \textit{Mod-Noun} classifier seems to be fooled in some cases by instances in which a modifier comes at the end of a phrase (e.g., ``a lovely misty gray''). 
% It could be relevant to what was revealed in \cite{sinha2021masked,alleman2021syntactic} that distributional information largely dominates the way that masked language modeling understands and encodes language.
For \textit{Verb-Ob}, the classifier seems at times to be fooled by grammatical sentences ending with a verb, or by fronting of prepositional phrases. 
% The truth that the encoders show constant patterns here, might be an explanation to the reason that Verb-Ob is the easiest anomaly type to be perceptible---certain properties might be relatively superficial to observe and encode with.
For \textit{SubN-ObN}, the false positives often involve nouns that are likely uncommon as subjects, such as ``bases''. All of these patterns suggest that to some extent, the anomaly-specific cues that the classifiers detect are closely tied to the particulars of our perturbations---some of which may constitute artifacts---and in some cases, they raise the question of whether classifiers can succeed on these tasks based on fairly superficial word position cues, rather than syntax per se. We follow up on this question in the following section.

\begin{figure}[t!]
 \centering
\includegraphics[width=0.48\textwidth]{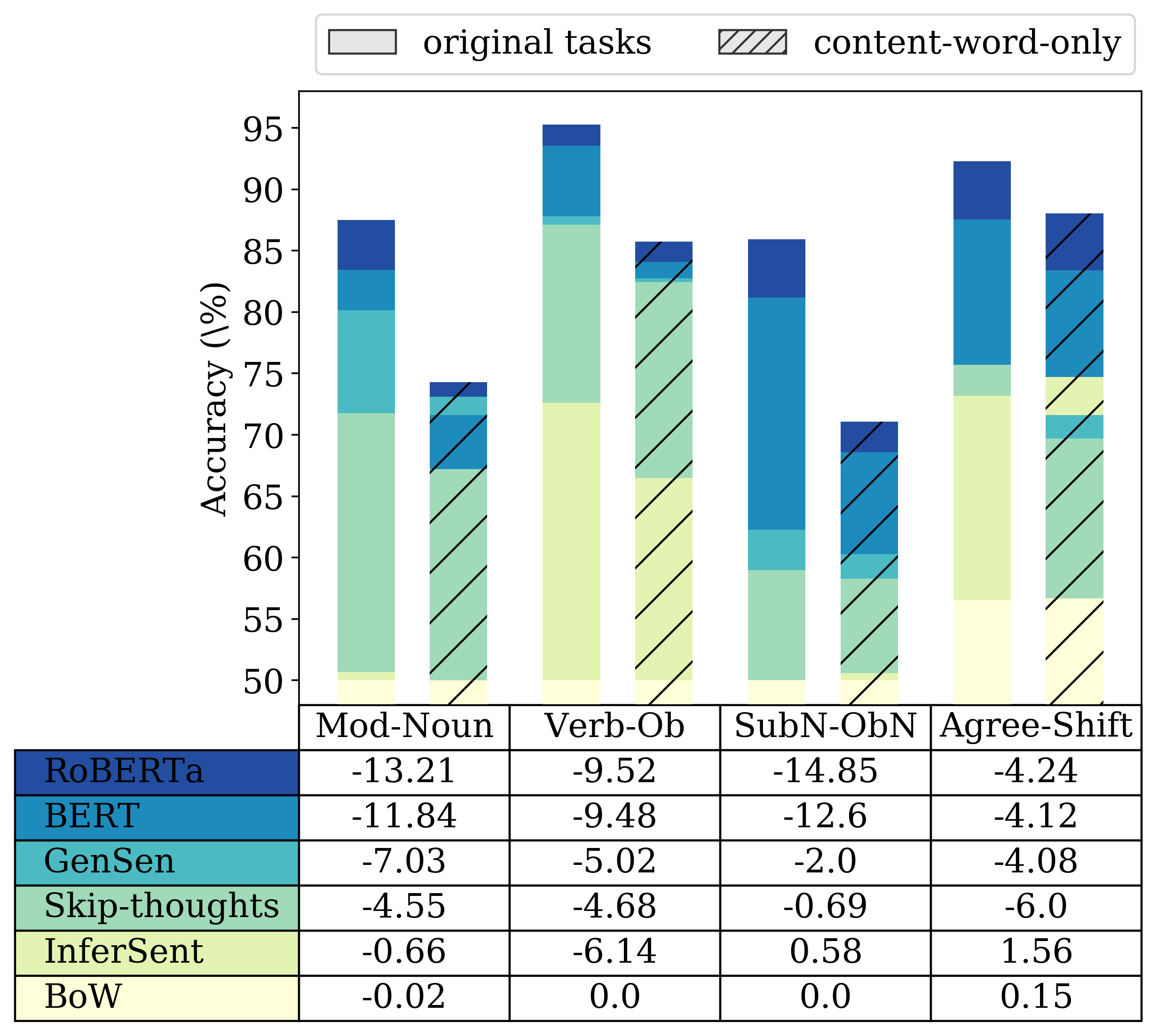}
 \caption{Performance of encoders on original tasks vs. ``content-word-only'' tasks (function words removed). Numbers in embedded table show change in accuracy from original to content-word-only setting. 
%  White bars show original results, and shaded bars show non-syntactic results. 
%  \ake{make larger}
%  Blank filled bars show original task results, which are the same as that shown in Fig. \ref{fig:original_task_only}.
%  Pattern-filled bars show non-syntactic task results. Different colors represent different encoding methods. 
%  Non-transparent bars show results on MLP classifier, while transparent bars show the results on LR classifier. All error bars represent 95\% binomial confidence intervals. 
%  Dotted horizontal line indicates chance-level accuracy. 
 }
 \label{fig:non_syntactic_vs_original}
\end{figure}

\subsection{Role of content word order}

To explore the possibility that classifiers may be succeeding in anomaly detection based on word position cues alone, rather than details of syntactic structure, we run a follow-up test using content-word-only versions of our probes. This serves as a test of how well the models can get by with coarser-grained information about content words positions.

Fig.~\ref{fig:non_syntactic_vs_original} shows anomaly detection performance when embeddings reflect only content words, as compared to the original anomaly detection performance. We see that across tasks, anomaly detection performance of Skip-thoughts, GenSen, BERT, and RoBERTa are all reduced as a result of the loss of function words.
% \footnote{As for InferSent, we see that accuracy is comparable with or without function words---even slightly worse with intact functional syntactic information on certain tasks. The model may simply be treating such functional information as noise.} 
BERT and RoBERTa in particular show substantial losses for the three reordering tasks, indicating that these models benefit significantly from function words for encoding the information that supports detection of these anomalies. It is also worth noting, however, that the models do retain a non-trivial portion of their original accuracy even with function words absent, supporting the idea that to some extent these perturbations can be detected through coarser position information rather than fine-grained syntax.\footnote{The notion that coarser position information alone can contribute non-trivially to anomaly identification is further supported by testing on \textit{normal} sentences in \textit{Subj-Num} when training on the content-word-only setting; for these results, see Appendix Table \ref{tab:wrongly_labeled_rate}.}
% \blue{shall we add this?} \ake{do you give detail about what the Table means, in the appendix?}}
This is an observation worth keeping in mind, particularly when interpreting anomaly detection as evidence of syntactic encoding~\cite[e.g.][]{Conneau18}.

\section{Discussion}

In the experiments above, we have taken a closer look at the nature of syntactic anomaly encoding in sentence embeddings. Using fine-grained variation in types of syntactic anomalies, we show differences in patterns of anomaly detection across encoders, suggesting corresponding differences in the types of anomaly information encoded by these models. 
While the margins of difference in anomaly-specific sensitivity are less dramatic between small RNN-based models and larger transformer models,
when we examine the generality of the detected anomaly signal, we find that only BERT and RoBERTa show signs of higher-level anomaly awareness, as evidenced by non-trivial transfer performance between anomalies.

What might be driving the anomaly encoding patterns indicated by our results? Explicit syntactic training does not appear to be necessary.
GenSen is the only model that includes an explicit syntactic component in its training (constituency parsing), which could help to explain that model's comparatively strong performance on the individual anomaly detection tasks. However, it is noteworthy that GenSen performs mostly on par with Skip-thoughts, which constitutes just one of GenSen's objectives, and which uses only prediction of adjacent sentences.
% Furthermore, while multiple of the RNN-based models show decent sensitivities on the individual anomaly tasks, only 
BERT and RoBERTa, the only models to show signs of more generalized anomaly encoding, have no explicit syntactic training at all. However, various findings have suggested that these types of models do develop syntactic sensitivity as a result of their more generalized training objectives~\cite{goldberg2019assessing,liu2019linguistic,alleman2021syntactic,tenney2019you}.
% This may shed light on what expensive model pre-training buys us, i.e., more generalized encoding but no salient improvement on anomaly-specific sensitivity. \ake{I'm not sure about that last sentence here, since the point being made is that syntactic objectives don't seem to be the critical factor in the models' success} 

We can imagine various ways in which objectives involving prediction of words in context, as used by BERT and RoBERTa, could encourage learning of a generalized notion of syntactic anomaly.
It may be the case that oddities occur together, in which case they could be mutually predictive and therefore of value for optimizing a prediction-based objective. More generally, anomalous sentences are likely identifiable as less probable, or more difficult to generate coherent predictions for. This relationship between anomaly and sentence probability raises a related question: Is this a problem for our conclusions here? Could models simply be identifying anomalous sentences as less probable, without any actual notion of syntactic anomaly?
In NLP models, assessment of text probabilities is closely related to assessment of text naturalness and acceptability.
For this reason, teasing apart general sensitivity to probability versus genuine awareness of syntactic grammaticality phenomena is a recurring challenge when testing syntactic knowledge in language models---and these things are similarly potentially entangled in our analyses here. To an extent this entanglement is inevitable and unproblematic: we necessarily expect syntactic anomalies to lower the probability of a string, and we can expect some awareness of syntactic anomaly to be important for assigning lower probability to an ungrammatical string. We can imagine, for instance, a situation in which a language model has no sensitivity to what constitutes good versus anomalous syntax, and thus assigns probability solely on the basis of word co-occurrence or other unrelated indicators of naturalness. In this sense, although it is not difficult to imagine how the close relationship between anomaly and sentence probability could be an explanation for findings that suggest anomaly awareness in these models, this does not change the fact that model representations may end up with genuine, detectable encoding of generalized anomaly information as a byproduct of probabilistic training---and this genuine anomaly encoding may be what we are detecting with our tests here. 
However, future work can examine further the relationship between syntactic anomaly and model perplexities, to explore whether these embeddings could show signs of anomaly sensitivity while in fact exclusively encoding confounding probabilistic information unrelated to syntactic anomaly.

Our content-word-only analysis provides one source of evidence on the relative importance of genuine syntactic sensitivity in success on our syntactic-related anomaly tasks.  
% follow-up analysis on models' reliance on finer-grained syntactic information, does provides some additional support for the notion that BERT and RoBERTa embeddings may indeed encode an abstract notion of syntactic oddity---in addition to more anomaly-specific cues.
% \orange{To what extent this abstract notion of syntactic anomaly awareness benefits from general-level syntactic sensitivity,  that these transformer models can non-trivially master of?}.
% Evidence has suggested that these transformer models learn non-trivial syntactic sensitivity  \cite{goldberg2019assessing,liu2019linguistic,alleman2021syntactic,tenney2019you}, but as for anomaly awareness per se, it is not yet clear whether such abstract syntactic knowledge directly benefits or guides the anomaly identification
This test aims to tease apart the extent to which success on our tasks requires processing of finer-grained syntactic information, versus the extent to which models can succeed based on more superficial content word position information. We find in this analysis that most encoders do benefit from the finer-grained syntactic information provided by function words, supporting an important role for more advanced syntactic sensitivity in these tasks---however, we also find that substantial proportions of the observed detection accuracy can indeed be achieved with content words alone. To the best of our knowledge, we are the first to report this finding. This leaves us with two key takeaways. First, to an extent there is good reason to believe that a reasonable amount of genuine syntactic sensitivity is involved in the highest levels of success on our anomaly tasks. Second, success on syntactic anomaly tasks can also be non-trivially inflated by use of more superficial cues.  That is to say, as usual, these datasets have a habit of enabling learning from simpler cues than are intended.
This takeaway highlights a need for caution in interpreting detection of reordering anomalies as evidence of deeper syntactic encoding per se~\cite{Conneau18}---especially in probing datasets that use naturally-occurring data without exerting controls for confounding cues.

While these results have shed light on potential encoding of generalized syntactic anomaly knowledge in pre-trained models, there are many further questions to pursue with respect to these models' handling and understanding of such anomalies. We will leave for future work the problem of understanding in greater detail how model training may contribute to encoding of a generalized awareness of anomalies in text, how a genuine notion of syntactic anomaly could be further disentangled from general probability sensitivity, and how one could exploit models' awareness of anomaly for improving model robustness on downstream pipelines. 
% Additionally, we would like to investigate the correlation between the perplexity degree and the magnitude of the anomaly signal we explored here in the future.

\section{Conclusion}
% \ake{Todo: update first paragraph of conclusion to more closely mirror introduction.} 
We have undertaken a direct study of anomaly encoding in sentence embeddings, finding impacts of hierarchical differences in anomaly type, but finding evidence of generalized anomaly encoding only in BERT and RoBERTa. Follow-up analyses support the conclusion that these embeddings encode a combination of generalized and anomaly-specific cues in these embeddings, with models appearing to leverage both finer-grained and coarser-grained information for anomaly detection. These results contribute to our understanding of the nature of encoding of linguistic input in embeddings from recent models. Future work can further explore the relationship between naturalness-oriented training and cultivation of abstract anomaly awareness, and how these insights can be leveraged for more robust and human-like processing of language inputs. 
% Other interesting future directions would be to apply our work on information-theoretic probing strategies \cite{voita2020information,lovering2021predicting}.
% , and to explore performance of more syntax-aware BERT variants \cite{bai2021syntax}.

\section*{Acknowledgments}

We would like to thank several anonymous reviewers for helpful comments and suggestions that have led to improvements on earlier versions of this paper. We would also like to thank the members of the University of Chicago CompLing Lab, for valuable discussions on this work.

% \clearpage

\bibliography{anthology,emnlp2021}
\bibliographystyle{acl_natbib}

% \bibliographystyle{acl_natbib}
% \bibliography{anthology,acl2021}

\appendix

\section{Appendix}\label{sec:appendix}

% We provide supplementary results about cross-lingual encoders, anomaly detection performances in terms of LR classifier,  follow-up analyses results (error analysis, and role of content words order), as well as
% % one-one transferring with Conneau et al. tasks, 
% the multi-one transferring results.
% Finally, we present descriptions about encoder models, external tasks, and some implementation details.
% % with both consistent training size and original training size 

\subsection{Cross-lingual encoders}
% \paragraph{Cross-lingual Results}

Our probing tasks focus on anomalies defined relative to English syntax---but of course grammatical properties vary from language to language. 
Some of our perturbations produce constructions that are grammatical in other languages.
% To explore how encoding of anomaly information is impacted by learning of a mapping between grammatical systems of distinct languages, we have included results from multiple cross-lingual variants of the Universal Sentence Encoder. 
% As we have noted above, all three of the Universal Sentence Encoder models for the most part show lower sensitivity to these anomalies than the other encoders. However, 
We compare Universal Sentence Encoder (Cer et al., 2018) in the variants 
% ~\cite{nmt_encoder}
of both monolingual and cross-lingual (Chidambaram et al., 2019)
% \cite{nmt_encoder_cross_lingual} 
trained models---Multi-task en-en, Multi-task en-fr, and Multi-task en-de.
This allows us to examine impacts of cross-lingual learning, between English and different target languages, on anomaly sensitivity. 
% \ake{wording okay?}
\footnote{We do not discuss this cross-lingual results in the main paper due to that the differences between mono-lingual encoder and its cross-lingual variants are small. But we do think the constant patterns behind the differences here are inspiring.}

From Table \ref{tab: cross-lingual-results}, we see that the two cross-lingual encoders (Multi-task en-fr and Multi-task en-de) do show slightly stronger anomaly detection relative to the monolingual model (Multi-task en-en) on \textit{Mod-Noun}, \textit{Verb-Ob}, and \textit{Agree-Shift}, while having similar accuracy on \textit{SubN-ObN}.
This suggests that to accomplish the cross-lingual mapping from English to French or English to German, these models may carry out somewhat more explicit encoding of syntactic ordering information, as well as morphological agreement information,
resulting in encoded embeddings being more sensitive to corresponding anomalies relative to the monolingual model.
As we have discussed in the main paper, anomaly detection in the \textit{SubN-ObN} task likely involves understanding extra world knowledge, so it is perhaps not surprising that the cross-lingual component does not provide a boost in sensitivity on that task. 
For the most part, we find that 
% the two cross-lingual models pattern together, suggesting that 
the difference between the English-to-French and English-to-German mapping does not significantly impact encoding of the tested anomaly types.

% \usepackage{graphicx}
\begin{table}[htbp]
\centering
\resizebox{0.5\textwidth}{!}{%
\begin{tabular}{l|l|l|l}
\toprule
   (accuracy \%)         & Multi-task en-en & Multi-task en-fr & Multi-task en-de \\ \midrule
\textit{Mod-Noun}    & 54.858           & 57.539           & 59.109           \\ \hline
\textit{Verb-Ob}     & 63.204           & 66.188           & 66.345           \\ \hline
\textit{SubN-ObN}    & 56.91            & 56.372           & 56.641           \\ \hline
\textit{Agree-Shift} & 56.125           & 61.746           & 61.33            \\ \bottomrule
\end{tabular}%
}
\caption{Anomaly detection results on mono-lingual encoder vs. its cross-lingual variants. (MLP)
}
\label{tab: cross-lingual-results}
\end{table}

As for the transferability of the anomaly encoding (especially on the multi-one transfer, see Table \ref{tab:multi_task_new_test_on_our_task_cosis_size}),
we observe non-trivial performance improvement in the multi-task setting for the Multi-task en-en model, relative to the cross-lingual variants. This suggests that Multi-task en-en may result in a somewhat more generalized anomaly encoding, while cross-lingual variants are more sensitive to properties of individual anomalies.

\subsection{Logistic regression results}

\begin{table}[htbp]
\resizebox{0.5\textwidth}{!}{%
\begin{tabular}{l|l|l|l|l|l|l}
\toprule
            & \multicolumn{6}{c}{LR}                                        \\ \midrule
            & BoW    & InferSent & \makecell[l]{Skip-\\thoughts} & GenSen & BERT   & Roberta \\ \hline
\textit{Mod-Noun}    & 50     & 57.595    & 69.329        & 79.459 & 82.903 & 87.189  \\ \hline
\textit{Verb-Ob}     & 50     & 72.078    & 85.001        & 87.469 & 93.033 & 95.187  \\ \hline
\textit{SubN-ObN}    & 50     & 56.451    & 57.774        & 61.723 & 80.48  & 85.484  \\ \hline
\textit{Agree-Shift} & 55.34  & 72.403    & 72.93         & 72.538 & 83.778 & 91.676  \\ \toprule
            & \multicolumn{6}{c}{MLP}                                       \\ \midrule
            & BoW    & InferSent & \makecell[l]{Skip-\\thoughts} & GenSen & BERT   & Roberta \\ \hline
\textit{Mod-Noun}   & 50     & 50.662    & 71.764        & 80.132 & 83.453 & 87.514  \\ \hline
\textit{Verb-Ob}     & 50     & 72.605    & 87.133        & 87.794 & 93.572 & 95.277  \\ \hline
\textit{SubN-ObN}    & 50     & 50        & 58.975        & 62.284 & 81.176 & 85.943  \\ \hline
\textit{Agree-Shift} & 56.518 & 73.166    & 75.701        & 74.736 & 87.536 & 92.282  \\ \bottomrule
\end{tabular}%
}
\caption{Results (accuracy \%) on original anomaly detection tasks, comparing between LR and MLP classifiers. The MLP results are the same as what has been shown in Fig. 1 in the main body.}
\label{tab:LR-results}
\end{table}

As seen in Table \ref{tab:LR-results}, the results suggest that, for the most part, training with LR yields comparable performance to that with MLP, consistent with the findings of~\citet{Conneau18}.\footnote{Since we observe performance of LR to be mostly on par with one-hidden-layer MLP, we expect benefits of exploration with further classifier complexity to be limited.}
We do find, however, that the LR classifier has higher accuracy for InferSent on \textit{Mod-Noun} and \textit{SubN-ObN}. This suggests that, to the extent that these anomalies are encoded (perhaps only weakly) in InferSent, they may in fact be better suited to linear extraction.
% \footnote{We also see that when training on the \textit{Mod-Noun} task with larger training size, performance of the MLP increases to approach that of LR. Based on this, we infer that 
% InferSent may need more training data to extract certain types of anomalous information with non-linear classifier.}
For the task of \textit{Agree-Shift}, most encoders show large improvements on MLP over LR, 
suggesting that morphological agreement anomalies are less conducive to linear extraction.

\begin{table*}[t!]
\centering
\resizebox{\textwidth}{!}{%
% \normalsize
\begin{tabular}{@{}l|l@{}}
\toprule

        & \multicolumn{1}{c}{\cellcolor[HTML]{C0C0C0}Exclusive in \textit{Mod-Noun}}                                                                                                                                                                                               
%         \\ \midrule
% GenSen  & \begin{tabular}[c]{@{}l@{}}The automaton 's attacks were straightforward and simple , but as the \textbf{minutes passed} it began throwing more combinations and feints .\\ Anyone sitting with us would never have known there was \textbf{anything weird between us} .\end{tabular} 
\\ \midrule
BERT    & \begin{tabular}[c]{@{}l@{}}
\makecell[l]{The buildings here were all \textbf{a lovely misty gray} , which gave them a dreamlike quality .}\\ There are terrible slums in \textbf{London Daisy} , places you 'd never want to visit .\end{tabular}                                                    \\ \midrule
RoBERTa & \begin{tabular}[c]{@{}l@{}}Suddenly , his senses sharpened and he felt \textbf{less inebriated} .\\ All the charts are in \textbf{drawers below the table} .\end{tabular}      
\\ \midrule
{\cellcolor[HTML]{C0C0C0}observation}
% {\cellcolor[HTML]{C0C0C0}cues}
& 
% \cellcolor[gray]{0.95}
\makecell[l]{
Most of the samples involve a construction which is ``seemingly a noun followed by a modifier format''.\\
For BERT, the samples seem to involve multiple adjectives in a row, where the final word is more frequently to be an adjective\\ generally, and is more clear following the \textit{Mod-Noun}-specific detection rules.  \\
% For GenSen and RoBERTa, 
For RoBERTa, e.g., “drawers below the table”, “drawers” actually belongs to another prepositional phrase “in drawers” which is parallel\\ to the followed by prepositional phrase “below the table”.}

\\ \midrule
        & \multicolumn{1}{c|}{\cellcolor[HTML]{C0C0C0}Exclusive in \textit{Verb-Ob}}                                                                                                                                                                                                
%         \\ \midrule
% GenSen  & \begin{tabular}[c]{@{}l@{}}\textbf{In the walls} were holes .\\ And fresh troops \textbf{kept coming} .\end{tabular}                                                                                                                                                                 
\\ \midrule
BERT    & \begin{tabular}[c]{@{}l@{}}Slowly , the gatehouse \textbf{rose} .\\ \textbf{Through their windows} , thick candles spread throughout flickered softly .\end{tabular}                                                                                                                  \\ \midrule
RoBERTa & \begin{tabular}[c]{@{}l@{}}\textbf{A satisfactory rate of exchange} I \textbf{feel} .\\ The entire column stretched back almost as far as the eye could \textbf{see} .\end{tabular}         

\\ \midrule
{\cellcolor[HTML]{C0C0C0}observation}  & 
\makecell[l]{A clear pattern across both encoders: the error samples involve fronting such as prepositional phrase-fronting or object-fronting, or involve\\ constructions end with a verb/verb phrase (thus not with a standard SVO structure).}

\\ \midrule
        & \multicolumn{1}{c}{\cellcolor[HTML]{C0C0C0}Exclusive in \textit{SubN-ObN}}                                                                                                                                                                                             
%         \\ \midrule
% GenSen  & \begin{tabular}[c]{@{}l@{}}Her \textbf{lipstick} is completely smudged , so that means she must 've snuck off and met with Matt .\\ The entry \textbf{deadline} is May fifteenth .\end{tabular}  
\\ \midrule
BERT    & \begin{tabular}[c]{@{}l@{}}The \textbf{object} changed as he spoke .\\ 
\makecell[l]{The \textbf{bases} were wide , and as the buildings climbed into the sky , they became narrower and branched off to connect to other buildings .}
\end{tabular}                                                  \\ \midrule
RoBERTa & \begin{tabular}[c]{@{}l@{}}That \textbf{joint} will help you sleep .\\ 
\makecell[l]{The stealth \textbf{assassin} never belonged , but the reason will shatter his every conviction .}\end{tabular}   

\\ \midrule
{\cellcolor[HTML]{C0C0C0}observation}  & 
\makecell[l]{For both encoders, the subject word of the sampled sentence is always an uncommon subject word.
% , while for GenSen, quite common.
% When focusing on the S-V-O structure, GenSen examples are more of acceptable when swapping subject and object.
}

\\ \midrule
        & \multicolumn{1}{c}{\cellcolor[HTML]{C0C0C0}Exclusive in \textit{Agree-Shift}}                                                                                                                                                                            
%         \\ \midrule
% GenSen  & \begin{tabular}[c]{@{}l@{}}A large blackened crater \textbf{opened} a few yards away .\\ The Prince 's adventures \textbf{continue} in THE RAVEN , coming in February from Penguin !\end{tabular}      
\\ \midrule
BERT    & \begin{tabular}[c]{@{}l@{}}Not even the vendors who \textbf{stood} at their little shops or at their carts and called out their specials cared that I was there .\\ The Tiger Man , still awake , \textbf{regarded} her with groggy eyes .\end{tabular}                               \\ \midrule
RoBERTa & \begin{tabular}[c]{@{}l@{}}My brows \textbf{went} up .\\ A humorless laugh \textbf{escaped} his mouth and all I could do was stand mute , my heart breaking .\end{tabular}

\\ \midrule
{\cellcolor[HTML]{C0C0C0}observation}  & 
\makecell[l]{Almost all of the error samples are with past tense main verb, across both encoders. \\
% GenSen, is the only encoder here occasionally wrongly labels sentences with present tense main verb.
}

\\ \bottomrule
\end{tabular}%
}
\caption{Sampled examples for error analysis, along with some basic observed patterns. We list sampled typical examples for which the sentences are false positives exclusively in each of our four tasks.
% , as well as when the sentences are false positives for all of three reordering tasks. 
% The top three encoders in the anomaly detection in general: GenSen, BERT, RoBERTa are considered in terms of each case. 
The bold text highlights words or constructions that possibly relate to what we think as cues that trigger our pre-trained classifier to predict the whole sentence as \textit{perturbed}.
% based on some implicit regularities the classifier learns/follows.
% \blue{TODO: move to appendix}
% Transfer tasks trained on Conneau et al. tasks. The columns printed with gray background list the test results on the training tasks themselves. (These are comparable, though not identical, to the results reported in the original paper, with differences likely due to changes in data split and hyperparameter settings.)
% Other columns list the performances of transfer tasks.
}
\label{tab:error_analysis_exclusive}
\end{table*}

\subsection{Error Analysis and the Role of Content Word Order}

We show the error analysis results on the out-of-sample \textit{Subj-Num} data in this part.
Table \ref{tab:wrongly_labeled_rate} shows the error rate in terms of each training task (original and content-word-only), with the top two strongest encoders within our investigation.
Table \ref{tab:error_analysis_exclusive} lists sampled false positives identified exclusively by each classifier, along with corresponding initial observations.

% \usepackage{graphicx}
\begin{table}[!ht]
\centering
\resizebox{0.48\textwidth}{!}{%
\begin{tabular}{l|l|l|l|l}
\toprule
           & \multicolumn{4}{c}{original}   \\ 
\midrule
train task & \textit{Mod-Noun} & \textit{Verb-Ob} & \textit{SubN-ObN} & \textit{Agree-Shift} \\ 
\hline
BERT       & 4.9\%             & 2.61\%           & 4.81\%            & 31.98\%               \\ \hline
RoBERTa    & 7.73\%            & 3.6\%            & 7.83\%           & 22.13\%              \\ \toprule
           & \multicolumn{4}{c}{content-word-only} \\ \midrule
train task & \textit{Mod-Noun} & \textit{Verb-Ob} & \textit{SubN-ObN} & \textit{Agree-Shift} \\ 
\hline
BERT       & 1.82\%            & 3.75\%           & 2.0\%             & 22.86\%              \\ 
\hline
RoBERTa    & 0.96\%            & 1.64\%           & 1.54\%            & 4.92\%               \\ 
\bottomrule
\end{tabular}%
}
\caption{Error rate on \textit{Subj-Num} data, with a total size of 10,000 sentences (test set). The top rows show the results when the training on original tasks, while the bottom rows show the results when training on content-word-only tasks.}
\label{tab:wrongly_labeled_rate}
\end{table}

As we see in the main paper, BERT and RoBERTa show non-trivial benefits from functional information for improving overall anomaly sensitivity, but the content-word-only setting can account for a substantial proportion of contribution for detecting the anomalies.
This is roughly consistent with what we found in Table \ref{tab:wrongly_labeled_rate} that training from content word order only can lead to relatively lower error rate for most cases on normal sentences.

\begin{table*}[thbp]
% \caption{Transfer results of multi-one transferring with consistent training size. The amount of training size of multi-task training is consistent with one-one transferring. The columns of $\Delta$ show how much the multi-one transferring improves or drops from the best one-one result. The improvements (positive $\Delta$ values) are bolded. }
% \label{tab:multi_task_new_consis_size}
\scriptsize
\centering
\newcommand{\tabincell}[2]{\begin{tabular}{@{}#1@{}}#2\end{tabular}}
% \subtable[Transfer tasks trained on multi-one transferring among our generated tasks.]{
\resizebox{0.85\textwidth}{!}{
\begin{tabular}{c|cc|cc|cc|cc}
\toprule
\multirow{3}*{Encoder/Train task} & \multicolumn{2}{c|}{\tabincell{c}{multi-task \\ (\textit{Verb-Ob} $+$ \\ \textit{SubN-ObN} $+$ \\ \textit{Agree-Shift})}}  & \multicolumn{2}{c|}{\tabincell{c}{multi-task \\ (\textit{Mod-Noun} $+$ \\ \textit{SubN-ObN} $+$ \\ \textit{Agree-Shift})}}  & \multicolumn{2}{c|}{\tabincell{c}{multi-task \\ (\textit{Mod-Noun} $+$ \\ \textit{Verb-Ob} $+$ \\ \textit{Agree-Shift})}} & \multicolumn{2}{c}{\tabincell{c}{multi-task \\ (\textit{Mod-Noun} $+$ \\ \textit{Verb-Ob} $+$ \\ \textit{SubN-ObN})}}  \\
 & \textit{Mod-Noun} & $\Delta$ & \textit{Verb-Ob} & $\Delta$ & \textit{SubN-ObN} & $\Delta$ & \textit{Agree-Shift} & $\Delta$ \\
 \midrule
%  \multirow{2}{*}{BoW} & LR & 50.0 & 0.0 & 50.0 & 0.0 & 50.0 & 0.0 & 48.857 & -1.143 \\
BoW  & 50.0 & 0.0 & 50.0 & 0.0 & 50.0 & 0.0 & 50.0 & 0.0 \\
 \hline
% \multirow{2}{*}{InferSent} & LR & 50.359 & -0.909 & 54.273 & -0.606 & 51.189 & -0.213 & 54.639 & -0.045 \\
InferSent & 50.011 & \textbf{0.011} & 54.61 & \textbf{0.012} & 51.369 & \textbf{0.92} & 54.337 & -0.1 \\
 \hline
% \multirow{2}{*}{Skip-thoughts} & LR & 53.119 & -0.695 & 61.934 & \textbf{2.277} & 51.178 & -1.133 & 54.46 & -0.09 \\
Skip-thoughts & 53.298 & -0.83 & 63.022 & \textbf{6.001} & 51.357 & -0.46 & 55.289 & \textbf{0.034} \\
 \hline
% \multirow{2}{*}{GenSen} & LR & 55.093 & -2.266 & 59.062 & \textbf{0.796} & 53.007 & -0.37 & 54.706 & \textbf{1.109} \\
GenSen & 55.531 & -1.772 & 60.61 & \textbf{2.456} & 52.984 & -0.213 & 54.729 & \textbf{1.401}  \\
\hline
% \multirow{2}{*}{BERT} & LR & 72.908 & -1.122 & 85.913 & -0.101 & 71.236 & -0.976 & 71.156 & \textbf{1.075} \\
BERT  & 73.996 & -0.101 & 83.614 & -0.919 & 72.616 & \textbf{0.09} & 71.257 & \textbf{0.941} \\
\hline
% \multirow{2}{*}{Multi-task en-en} & LR & 53.332 & \textbf{0.281} & 54.071 & -1.963 & 53.096 & \textbf{0.493} & 50.594 & -0.739 \\
Multi-task en-en  & 53.455 & \textbf{0.123} & 54.587 & -2.804 & 52.984 & -0.247 & 51.154 & \textbf{0.258} \\
\hline
% \multirow{2}{*}{Multi-task en-fr} & LR & 52.289 & -0.426 & 54.06 &\textbf{ 0.404} & 52.053 & -0.998 & 52.678 & \textbf{0.056} \\
Multi-task en-fr & 52.165 & -0.797 & 53.41 & -0.28 & 52.244 & -1.054 & 52.824 & -0.258 \\
\hline
% \multirow{2}{*}{Multi-tasken-de} & LR & 52.221 & \textbf{0.224} & 53.454 & -0.595 & 52.524 & -0.157 & 52.129 & -0.796 \\
Multi-task en-de & 52.008 & -0.381 & 54.419 &\textbf{0.179} & 52.333 & -0.73 & 52.387 & -0.863 \\
 \bottomrule
\end{tabular}
}
\caption{
% Transfer tasks trained on multi-one transferring among our generated tasks. 
Transfer results of multi-one transferring among our generated tasks with consistent training size. The amount of training size of multi-task training is consistent with one-one transferring. The columns of $\Delta$ show how much the multi-one transferring improves or drops from the best one-one result. The improvements (positive $\Delta$ values) are bolded. }
\label{tab:multi_task_new_test_on_our_task_cosis_size}
\end{table*}

\begin{table*}[!ht]
% \caption{Transfer results of multi-one transferring with consistent training size. The amount of training size of multi-task training is consistent with one-one transferring. The columns of $\Delta$ show how much the multi-one transferring improves or drops from the best one-one result. The improvements (positive $\Delta$ values) are bolded. }
% \label{tab:multi_task_new_consis_size}
\scriptsize
\centering
\newcommand{\tabincell}[2]{\begin{tabular}{@{}#1@{}}#2\end{tabular}}
% \subtable[Transfer tasks trained on multi-one transferring among our generated tasks.]{
% \subtable[Transfer tasks jointly trained on multi-task learning with all of Conneau et al. tasks, tested on each of our generated tasks. ]{
\resizebox{0.8\textwidth}{!}{
\begin{tabular}{c|cc|cc|cc|cc}
\toprule
    \multirow{3}*{Encoder/Train task} & \multicolumn{8}{c}{multi-task (\textit{SOMO} $+$ \textit{BShift} $+$ \textit{CoordInv})} \\
    &  \multicolumn{2}{c|}{\textit{Mod-Noun}}  & \multicolumn{2}{c|}{\textit{Verb-Ob}}  & \multicolumn{2}{c|}{\textit{SubN-ObN}} & \multicolumn{2}{c}{\textit{Agree-Shift}}  \\
    & acc & $\Delta$ & acc & $\Delta$ & acc & $\Delta$ & acc & $\Delta$ \\ 
    \midrule
% Encoder & Classifier & train & Multi-task (SOMO + \textit{BShift} + \textit{CoordInv}) &  &  &  &  &  &  &  \\
%  &  & test & modifier-NN reorder &  $\Delta$ & Two NN reorder &  $\Delta$ & \textit{Verb-Ob} & $\Delta$ & VB\_transfer &  $\Delta$ \\
    % \multirow{2}{*}{BOW} & LR & 50.033 & -0.55 & 50.033 & -0.4 & 50.017 & -0.416 & 50.383 & \textbf{0.216} \\
BOW & 50.0 & 0.0 & 50.033 & \textbf{0.033} & 50.017 & \textbf{0.017} & 50.0 & 0.0 \\
 \hline
% \multirow{2}{*}{Infersent} & LR & 51.183 & -0.184 & 51.283 & -1.534 & 50.567 & -0.383 & 50.5 & -1.633 \\
Infersen  & 50.067 & -0.833 & 50.217 & -2.316 & 50.1 & -1.2 & 49.933 & -0.5 \\
 \hline
% \multirow{2}{*}{Skip-thoughts} & LR & 55.867 & -1.816 & 56.117 & -3.916 & 50.517 & -1.183 & 50.65 &\textbf{ 0.367} \\
Skip-thoughts  & 57.767 & -2.666 & 57.733 & -5.05 & 51.3 & -0.85 & 50.383 & -0.4 \\
 \hline
% \multirow{2}{*}{GenSen} & LR & 57.9 & -4.567 & 59.25 & -4 & 51.067 & -0.933 & 51.433 & -0.45 \\
GenSen  & 58.533 & -4.75 & 58.867 & -4.75 & 50.533 & -1.834 & 50.783 & -1.617  \\
 
  \hline
% \multirow{2}{*}{BERT} & LR & 69.333 & -3.684 & 72.8 & -5.4 & 68.917 & -0.966 & 62.583 & -1.967  \\
BERT  & 69.883 & -3.967 & 73.683 & -5.1 & 69.317 & -0.75 & 62.367 & -1.55 \\
 
 \hline
% \multirow{2}{*}{Multi-task en-en} & LR & 52.233 & -0.6 & 54.4 & \textbf{0.117} & 51.917 & -0.466 & 50.267 & -0.266 \\
Multi-task en-en  & 53.283 & \textbf{0.283} & 54.0 & -0.533 & 52.617 & \textbf{0.134} & 51.017 & \textbf{0.084} \\
 \hline
% \multirow{2}{*}{Multi-task en-fr} & LR & 51.267 & -0.216 & 52.283 & -4.65 & 51.367 & -0.75 & 51.017 & -0.116 \\
Multi-task en-fr  & 50.767 & -0.933 & 52.5 & -5.467 & 51.85 & -0.45 & 51.05 & \textbf{0.117} \\
 \hline
% \multirow{2}{*}{Multi-task en-de} & LR & 49.25 & -3.983 & 50.383 & -5.434 & 49.733 & -1.984 & 49.817 & -0.5 \\
Multi-task en-de  & 49.7 & -3.4 & 50.167 & -5.3 & 50.3 & -1.167 & 50.033 & -0.5 \\
\bottomrule
\end{tabular}}
\caption{
Transfer tasks jointly trained on multi-task learning with all of Conneau et al. tasks, tested on each of our generated tasks,
% Transfer results of multi-one transferring 
with consistent training size. 
% The amount of training size of multi-task training is consistent with one-one transferring. The columns of $\Delta$ show how much the multi-one transferring improves or drops from the best one-one result. The improvements (positive $\Delta$ values) are bolded. 
}
\label{tab:multi_task_new_conneau_generated_cosis_size}
% }
\end{table*}

\subsection{Transferring via Multi-Task Learning}

Tables~\ref{tab:multi_task_new_test_on_our_task_cosis_size}-\ref{tab:multi_task_new_conneau_generated_cosis_size} 
list the multi-one transferring results with consistent training size compared to that of one-one transferring. 
% \footnote{We also tried multi-one transferring results with the original training size on each training task, i.e., the amount of training size of multi-task training is three times of one-one transferring. The results from consistent size and from original size are mostly similar, while with less amount training data, the overall transfer performance benefiting from multi-anomaly learning slightly drops. For some cases, with consistent training size even yields better transfer generalization capacity than that with larger training data available.}
% Table \ref{tab:multi_task_new_test_on_our_task_whole_size} and Table \ref{tab:multi_task_new_conneau_generated_whole_size} list the multi-one transferring results with the original training size on each training task.
We show the multi-one transfer results along with the performance change relative to the best one-one results obtained on the current test task when training with any one of the joint training tasks. 
% The results from consistent size and from original size are mostly similar, while with less amount training data, the overall transfer performance benefiting from multi-anomaly learning slightly drops. For some cases, with consistent training size even yields better transfer generalization capacity than that with larger training data available.
Most of the multi-one transfer results show small amount of decrease from one-one transfer, suggesting that classifiers are still fitting to anomaly-specific properties that reduce transfer of anomaly detection. 
% BERT’s performance reductions for multi-one transfer are notably larger with the MLP classifier than with LR—further supporting the previous finding that the more generalized anomaly signal is better-suited for linear extraction.
% Multi-task en-en shows the most substantial improvement with multi-one transferring, not only relative to its cross-lingual variants, but among all encoders. 
% As we discuss above, the cross-lingual models may need to acquire somewhat stronger syntactic sensitivity, allowing them higher performance when detecting certain anomaly information. However, the pattern reversal observed here suggests that this extra sensitivity is specific to the characteristics of the corresponding perturbations, resulting in worse generalization in the transfer setting.

\subsection{Description of External Tasks for Transfer Training}

\paragraph{\textit{SOMO}}
\textit{SOMO} distinguishes whether a randomly picked noun or verb was replaced with another noun or verb in a sentence.
\paragraph{\textit{BShift}}
\textit{BShift} distinguishes whether two consecutive tokens within a sentence have been inverted.
\paragraph{\textit{CoordInv}}
\textit{CoordInv} distinguishes whether the order of two co-ordinated clausal conjoints within a sentence has been inverted.
\paragraph{\textit{CoLA}}
\textit{CoLA} tests detection of general linguistic acceptability in natural occurring corpus, using expert annotations by humans. 
% See Table \ref{tab:cola_data} for some representative unacceptable examples.

\subsection{Description of Encoders}
% \begin{itemize}
\paragraph{InferSent} InferSent is a sentence encoder optimized for natural language inference (NLI),
% with the Stanford Natural Language Inference (SNLI) dataset
mainly focusing on capturing semantic reasoning information for general use.
\paragraph{Skip-thoughts} Skip-thoughts is a sentence encoder framework trained on the Toronto BookCorpus, with an encoder-decoder architecture, to reconstruct sentences preceding and following an encoded sentence. 
    % \ake{I don't know that it makes sense to call this one unsupervised and InferSent supervised, since they are both using learning signals not involving the sentence representation task itself.}
\paragraph{GenSen} GenSen is a general-purpose sentence encoder trained via large-scale multi-task learning. 
Training objectives include Skip-thoughts, NLI, machine translation, and constituency parsing.
\paragraph{BERT} BERT is a deep bidirectional transformer model, pre-trained on tasks of masked language modeling (MLM) and next-sentence prediction (NSP). 
% To produce a sentence representation from BERT, we take the average of pre-trained embeddings from the last layer for all tokens of the sentence.
% , which has advanced the state-of-the-art in a wide variety of NLP tasks. 
% \footnote{We use BERT\textsubscript{LARGE}, uncased, with whole word masking. }
\paragraph{RoBERTa} RoBERTa is a variant of BERT, and outperforms BERT on a suite of downstream tasks. RoBERTa builds on BERT's MLM strategy, removing BERT’s NSP objective, with improved pre-training methodologies, such as dynamically masking.
\footnote{We use bert-large-uncased-whole-word-masking, and roberta-large. We take the average of fixed pre-trained embeddings from the last layer for all sentence tokens.
Pilot experiments show comparable performance between the average of all token embeddings versus the first/CLS token embedding.
}

\subsection{Implementation Details}
\paragraph{Hyperparameters}
Dropout rate is set to be $0.25$. Batch size is set to be 64. 
Early stopping is applied. The optimizer is Adam. The learning rate is explored within \{0.01, 0.001, 0.0001, 0.00001\}.
The MLP classifier has one hidden layer of 512 units. 

\paragraph{Probing data generation}
We use the premise sentences from the train, dev-matched, dev-mismatched datasets of MultiNLI, with repeats discarded according to the promptID.
\footnote{The following tools are used for our generation: nltk.tree \url{https://www.nltk.org/\_modules/nltk/tree.html}, and spaCy \url{https://spacy.io}. The transformation tool (along with extra rules) for \textit{Agree-Shift}: https://www.clips.uantwerpen.be/pages/pattern.}
\footnote{The ratio of plural verbs to single verbs (VBP/VBZ) in the original sentences is 1.044/1.}

We adopt an approach that we refer to as ``exhaustive'' perturbation: modifying all instances of a given structure within a sentence,
to ensure that sentences have internal structural consistency---e.g., a perturbed sentence in \textit{Mod-Noun} will not contain both ``modifier+noun'' and ``noun+modifier'' structures---thus avoiding inconsistency serving as an extra signal for detection.
\footnote{The average number of modifications per sentence is 1.69, 1.97, 1.0, and 1.97 for \textit{Mod-Noun},  \textit{Verb-Ob}, \textit{SubN-ObN},  and \textit{Agree-Shift}), respectively. Note that when we instead restrict perturbations to a single modification per sentence, we see that the same basic patterns across tasks are retained.}

For each task, we use training data of 71k sentences, and dev and test data of 8.9k sentences.

For the \textit{SubN-ObN} and the \textit{Mod-Noun} tasks,\footnote{For the task of \textit{Mod-Noun},
this could in particular happen with noun phrases involving noun-noun compounds.}
sometimes the case might arise that the resulting perturbed sentences are still normal or acceptable, but perhaps somewhat stranger or less probable to occur in the wild, e.g., “man bites dog”. However, this should be a rare case, as the original sentences are long enough to involve adequate context to distinguish normal from perturbed examples.

\end{document}